\documentclass{bmvc2k}

\title{Adapting Self-Supervised Representations to Multi-Domain Setups}

\addauthor{Neha Kalibhat}{nehamk@umd.edu}{1}
\addauthor{Sam Sharpe}{samuel.sharpe@capitalone.com}{2}
\addauthor{Jeremy Goodsitt}{jeremy.goodsitt@capitalone.com}{2}
\addauthor{Bayan Bruss}{bayan.bruss@capitalone.com}{2}
\addauthor{Soheil Feizi}{sfeizi@cs.umd.edu}{1}

\addinstitution{
 University of Maryland \\
 College Park, MD, USA \\ 
}
\addinstitution{
 CapitalOne Research \\
}

\runninghead{Kalibhat \etal}{Adapting SSL Representations using DDM}

\def\etal{\emph{et al}\bmvaOneDot}

\usepackage{graphicx}
\usepackage{amsmath}
\usepackage{amssymb}
\usepackage{booktabs}

\usepackage{amssymb}
\usepackage{mathtools}
\usepackage{amsthm}
\usepackage{multirow}
\usepackage{bbm}
\usepackage{subfigure}
\usepackage{wrapfig}

\newcommand{\bx}{\mathbf{x}}

\newcommand{\bc}{\mathbf{c}}

\newcommand{\bbR}{\mathbb{R}}

\newcommand{\bh}{\mathbf{h}}

\begin{document}

\maketitle

\begin{abstract}
Current state-of-the-art self-supervised approaches, are effective when trained on individual domains but show limited generalization on unseen domains. We observe that these models poorly generalize even when trained on a mixture of domains, making them unsuitable to be deployed under diverse real-world setups. We therefore propose a general-purpose, lightweight Domain Disentanglement Module (DDM) that can be plugged into any self-supervised encoder to effectively perform representation learning on multiple, diverse domains with or without shared classes. During pre-training according to a self-supervised loss, DDM enforces a disentanglement in the representation space by splitting it into a domain-variant and a domain-invariant portion. When domain labels are not available, DDM uses a robust clustering approach to discover pseudo-domains. We show that pre-training with DDM can show up to $3.5\%$ improvement in linear probing accuracy on state-of-the-art self-supervised models including SimCLR, MoCo, BYOL, DINO, SimSiam and Barlow Twins on multi-domain benchmarks including PACS, DomainNet and WILDS. Models trained with DDM show significantly improved generalization ($7.4\%$) to unseen domains compared to baselines. Therefore, DDM can efficiently adapt self-supervised encoders to provide high-quality, generalizable representations for diverse multi-domain data. 
\end{abstract}

\begin{figure}[h]
    \centering
    \includegraphics[width = \textwidth]{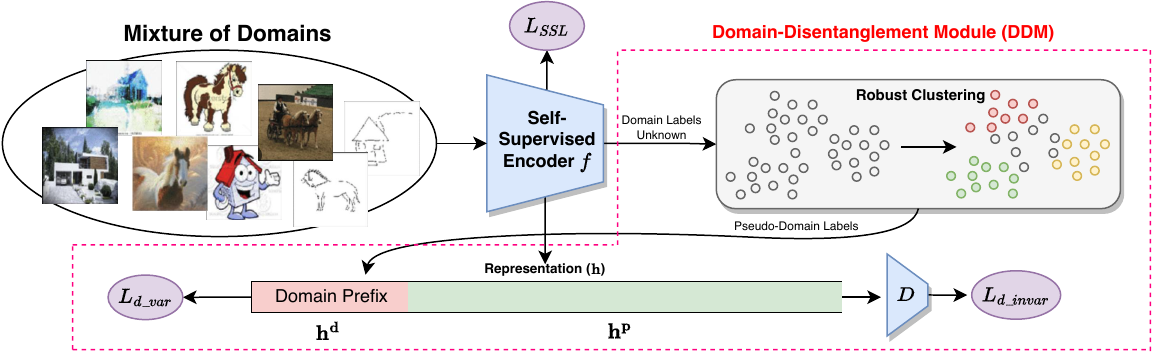}
    \caption{\textbf{Framework of our proposed Domain Disentanglement Module:} In our proposed DDM framework, the representation space ($\bh$) of any given self-supervised encoder is split into two portions, a domain prefix ($\bh^d$) and a domain-invariant ($\bh^p$) portion. Along with the self-supervised loss ($L_\textit{ssl}$), $\bh^d$ is trained to be distinguishable across domains ($L_\textit{d\_var}$) and $\bh^p$ is trained to be invariant to any domain information ($L_\textit{d\_invar}$). DDM also supports scenarios when domain labels are not available using \textit{robust clustering}, an iterative process that reduces outlier noise.}
    \label{fig:mdssl_framework}
\end{figure}

\section{Introduction}
\begin{figure}[h]
    \centering
    \subfigure{\includegraphics[width=0.32\textwidth]{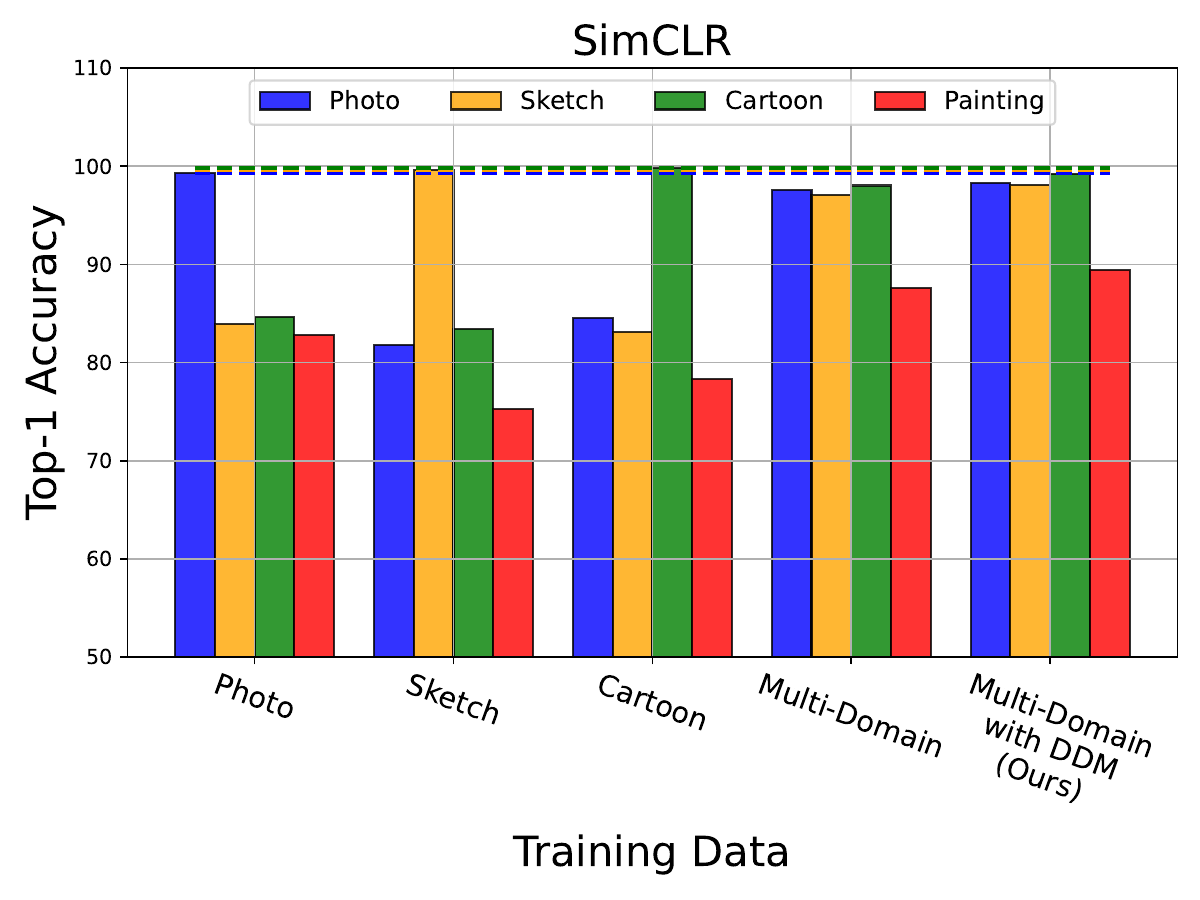}}
    \subfigure{\includegraphics[width=0.32\textwidth]{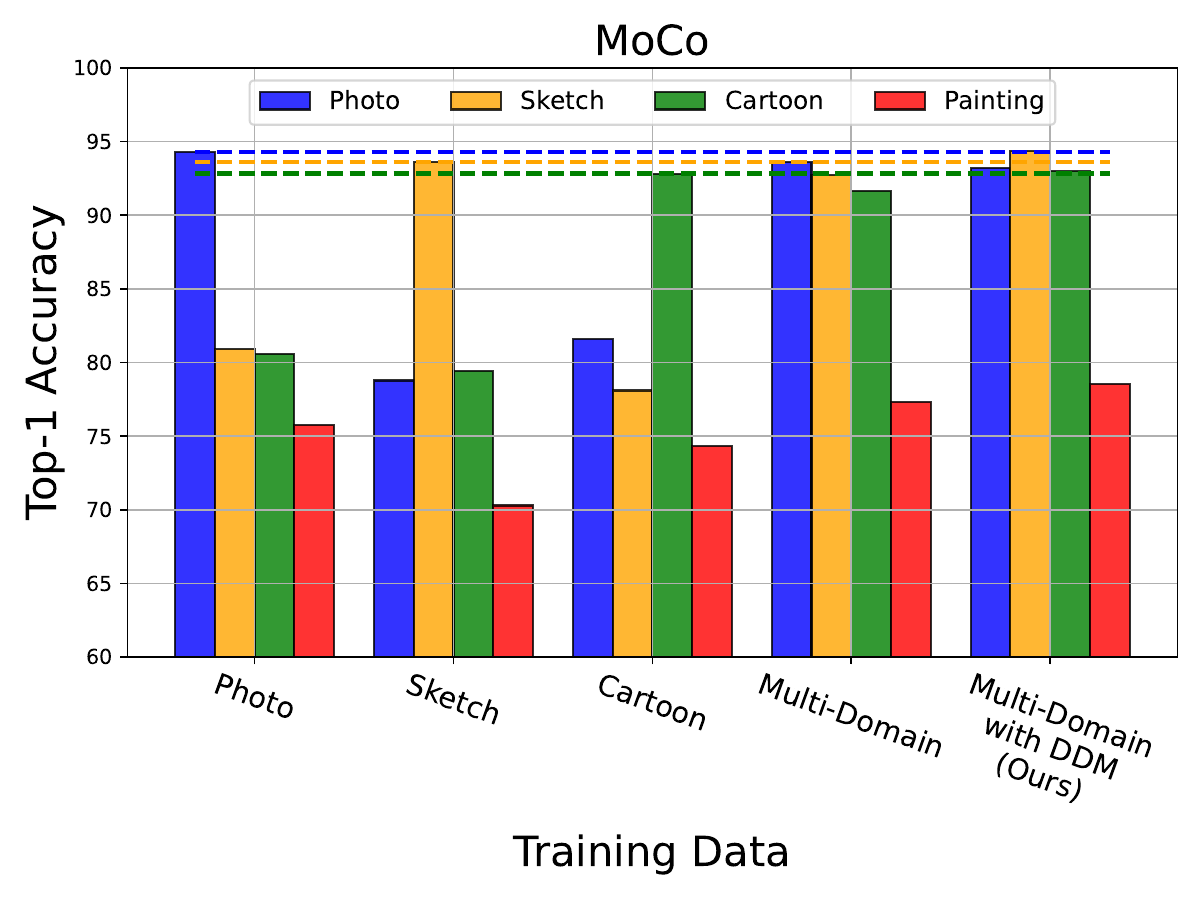}}
    \subfigure{\includegraphics[width=0.32\textwidth]{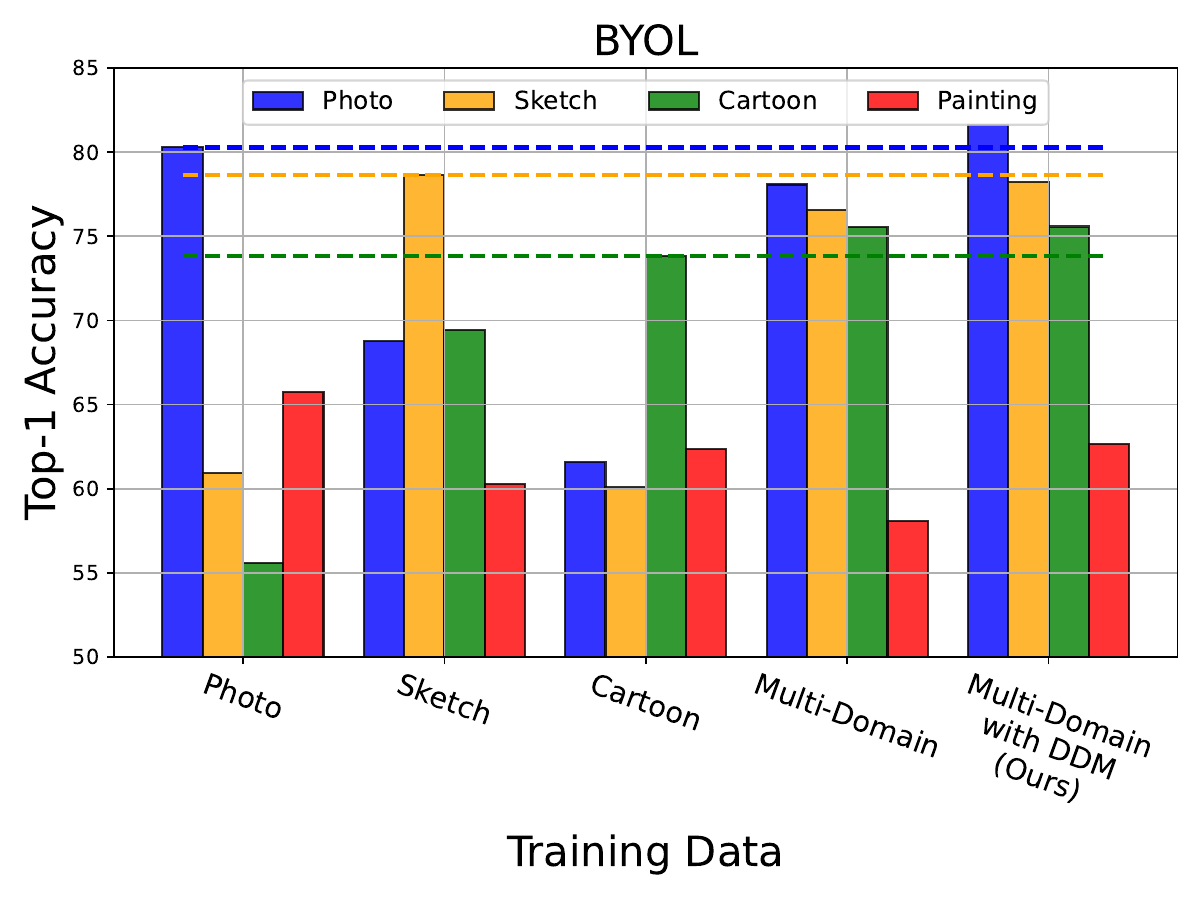}}
    \caption{\textbf{Self-supervised baselines under single and multi-domain setups:} We plot 3 SOTA self-supervised baselines, SimCLR, BYOL and MoCo, trained individually on PACS Photo, Sketch and Cartoon and on their mixture. We observe that both single-domain and mult-domain training generalizes poorly to unseen domains on all baselines. These baselines when pre-trained with DDM (our method), outperforms even single-domain baselines and shows significantly improved generalization to the unseen domains. }
    \label{fig:single_domain}
\end{figure}

Self-supervised learning \cite{simclr, swav, moco, byol, simsiam, deepcluster, khosla2020supervised, dino} has become a popular paradigm for unsupervised representation learning as it shows impressive results on downstream tasks. However, we find that current self-supervised models when trained on a single-domain show very poor generalizability to domain shifts. This can hinder their deployment in large scale real-world settings where data almost always comes from multiple diverse domains. We illustrate this issue in Figure \ref{fig:single_domain}, where we show that popular self-supervised models, SimCLR \cite{simclr}, MoCo \cite{moco} and BYOL \cite{byol}, trained on individual domains of PACS \cite{pacs} significantly under-perform on unseen domains. This means that a different self-supervised model needs to be trained for every new domain, which can add significant computational overheads given that training these models often require large batch sizes and a large number of training epochs \cite{simclr, moco, Wu_2018_CVPR}. 

One potential solution for self-supervised learning on multi-domain datasets is to train the models on the {\it union} of all input domains. We illustrate this in Figure \ref{fig:single_domain} were we plot the multi-domain training results for each baseline on a mixture of PACS Photo, Sketch and Cartoon. We observe that this solution may show improved performance on the training domains, however they do not match the single-domain baselines in all cases. Moreover, they show poor generalization to unseen domains (PACS Painting). In Section \ref{sec:multi_domain_ssl}, we study the representation space closely under multi-domain regimes to find that they can under perform compared to single-domain regimes because domain-related and content-related information overlap in the representation space, affecting their quality for instance classification.

To tackle these issues, we propose a {\bf Domain-Disentanglement Module (DDM)}, that can be plugged in to any self-supervised model during multi-domain training. With DDM, we enforce a disentanglement in the representation space where a domain prefix is trained to be distinguishable across domains and the remaining portion is trained to be \textit{domain-invariant} to produce better structured representations. This is achieved by minimizing the Wasserstein Distance \cite{arjovsky2017wasserstein} between the known and predicted domain label distributions. We also extend DDM to more realistic, entirely unsupervised multi-domain setups where domain labels are unknown. In such scenarios, we present a \textit{robust clustering} approach that iteratively reduces outlier noise and detects pseudo-domain-labels that are used in DDM.

By pre-training with DDM, we show that we can improve the generalization capability of various state-of-the-art self-supervised baselines including SimCLR \cite{simclr}, MoCo \cite{moco}, BYOL \cite{byol}, DINO \cite{dino}, SimSiam \cite{simsiam} and Barlow Twins \cite{barlowtwins}. We perform extensive experiments on generalization benchmarks including PACS \cite{pacs}, DomainNet \cite{peng2019moment} and WILDS \cite{koh2021wilds}. Upon linear probing on unseen domains, we observe an improvement of $6.1\%$ on PACS, $7.4\%$ on DomainNet and $5.9\%$ on WILDS using DDM. In summary, we propose a lightweight module called DDM which can be simply attached to any self-supervised encoder to enable training over multiple diverse domains to produce well-structured, generalizable representations (See Figure \ref{fig:mdssl_framework}).

\section{Related Work}
Building on the success of unsupervised learning techniques \cite{pmlr-v70-bojanowski17a, NIPS2014_07563a3f, YM.2020Self-labelling, NIPS2016_65fc52ed, Caron_2018_ECCV, Caron_2019_ICCV, pmlr-v97-huang19b}, self-supervised models have shown unprecedented capabilities when used in a range of downstream tasks. Among a number of self-supervised baselines, we focus on SimCLR \cite{simclr}, MoCo \cite{moco}, BYOL \cite{byol}, DINO \cite{dino}, SimSiam \cite{simsiam} and Barlow Twins \cite{barlowtwins}. These are joint-embedding self-supervised learning methods, which involve taking two augmented views of the same input and ensuring their representations are close using the same encoder or two encoders sharing the same weights. 

Extending these self-supervised methods to multiple diverse domains, other than ImageNet \cite{ILSVRC15}, is a relatively less explored topic \cite{10.1007/978-3-030-58574-7_43}. Existing approaches \cite{Kim_2021_ICCV, li2021invariant, kim2020crossdomain} use pre-trained encoders and assume few source labels for unsupervised domain adaption and domain generalization. \cite{sun2022opencon} uses available class information and novelty discovery to learn new samples in the wild. These works do not consider fully unsupervised multi-domain setups, where even domain label information is unavailable. \cite{Feng2019SelfSupervisedRL} assumes domain labels and uses mutual information to encode common invariant information and domain-specific information for each image. \cite{yang2022domain} uses multiple domain-specific decoders to reconstruct images according to their domains such that the encoder is domain-invariant. This method may not be scalable and is contingent upon the number of available domains. \cite{diul} proposes a contrastive method that selects negatives across domains to train invariant representations. Our method reports better numbers on the PACS dataset compared to these baselines. Our method also does not assume domain labels and can be flexibly applied on any self-supervised setup. 

In our paper, we focus on a general multi-domain setup with diverse related or unrelated domains, with and without shared classes, and evaluate on individual domain-specific tasks. We make it possible to efficiently pre-train a single encoder on any existing state-of-the-art self-supervised setup, over multiple domains, to significantly improve their generalizability. 


\begin{figure}
    \centering
    \subfigure{\includegraphics[width=0.3\textwidth]{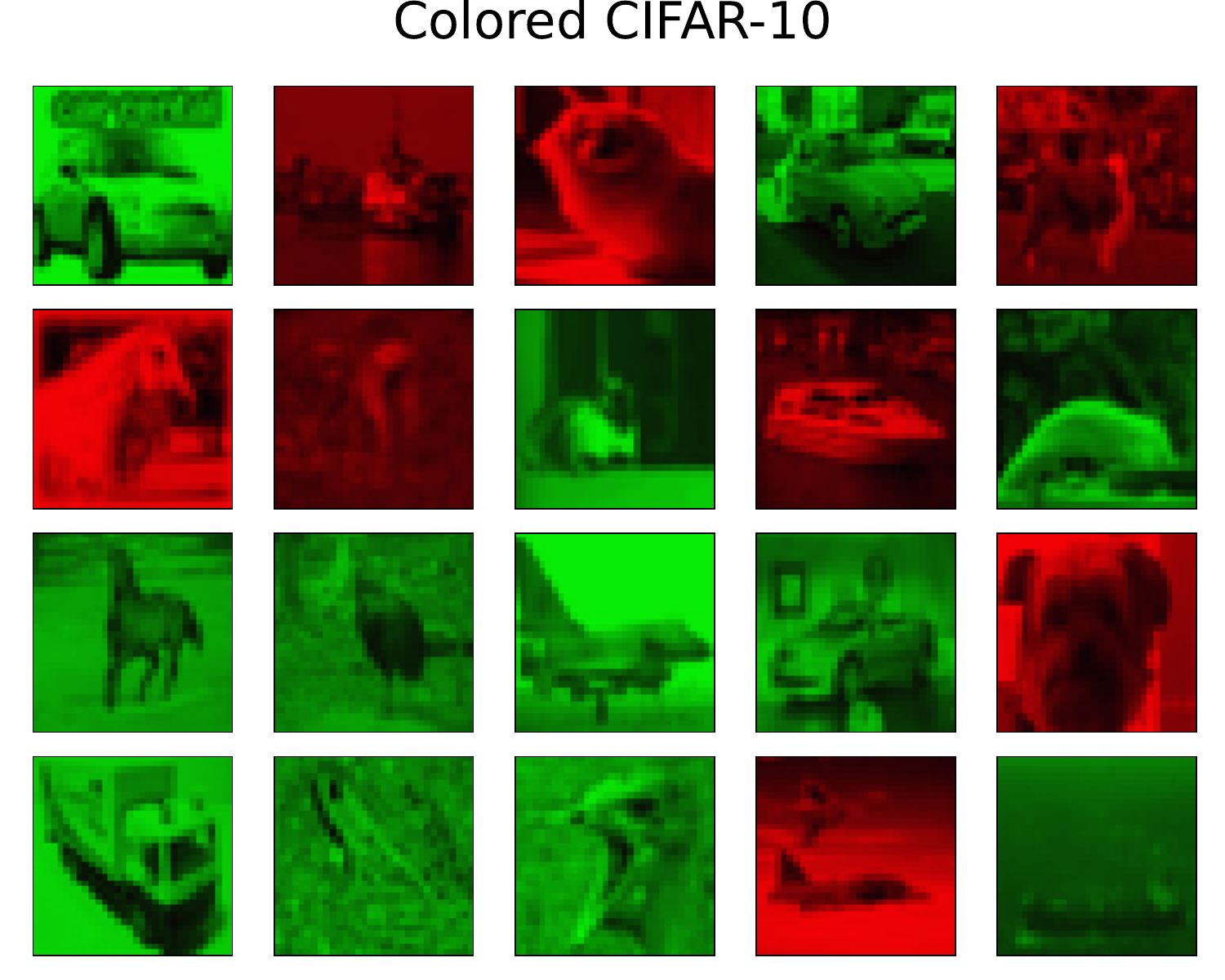}}
    \subfigure{\includegraphics[width=0.3\textwidth]{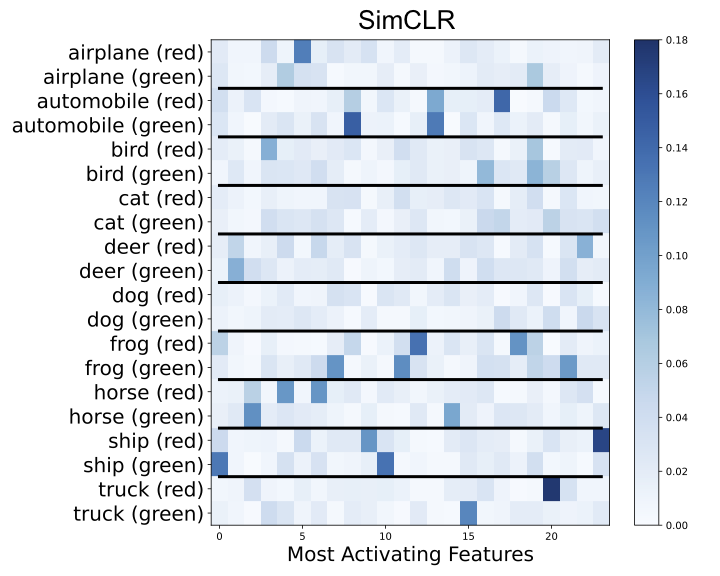}}
    \subfigure{\includegraphics[width=0.3\textwidth]{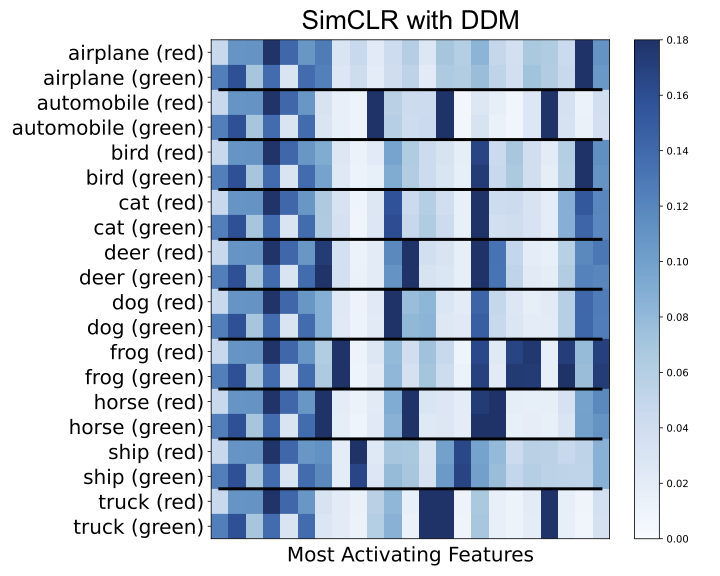}}
    \caption{\textbf{Visualizing Colored-CIFAR representations:} We prepare Colored-CIFAR, multi-domain version of CIFAR-10 \cite{cifar10} where the images are randomly colored red or green. We visualize the top activating features of the class-averaged representations of both domains. In the SimCLR baseline, we observe a clear difference in feature distribution within the same classes, across domains. DDM enables representations to have a shared domain-specific prefix while the remaining portion is domain-invariant and almost identical across classes. This structure significantly improves linear evaluation performance (See Figure \ref{fig:tsne_comp}).}
    \label{fig:most_act_feat}
\end{figure}
\section{Self-Supervised Models under Multi-domain Setups} \label{sec:multi_domain_ssl}
We observed in Figure \ref{fig:single_domain}, that state-of-the-art self-supervised learning methods like SimCLR \cite{simclr}, MoCo \cite{mocov2} and BYOL \cite{byol} show low transfer performance on unseen domains on both single-domain and multi-domain regimes. In this section, we take a closer look at the learned representation space under these regimes to explain this behavior. 

We first define some notations. Let us consider a self-supervised model with a base encoder $f(.)$. We apply data transformations and pass the input samples, $\bx_i \in \bbR^n$, through the base encoder to get self-supervised representations denoted by $f(\bx_i) = \bh_i \in \bbR^r$ where $r$ is the size of the representation space. 



\begin{figure}[h]
    \centering

    \scriptsize
    \setlength{\tabcolsep}{0.3pt}
    \begin{tabular}{cccc}
    \includegraphics[width=0.24\textwidth]{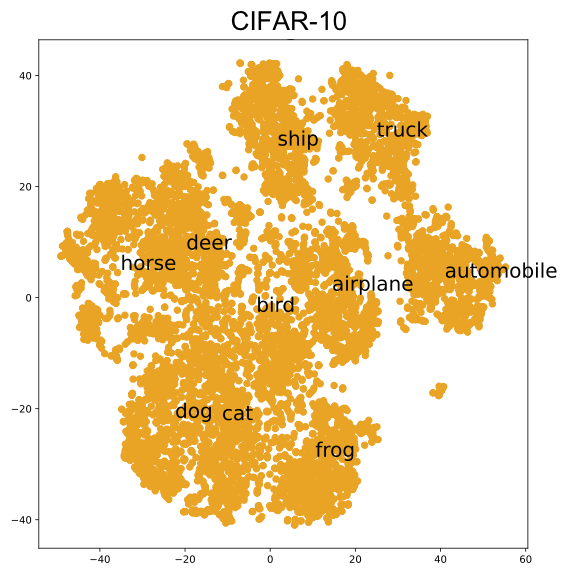}&
    \includegraphics[width=0.24\textwidth]{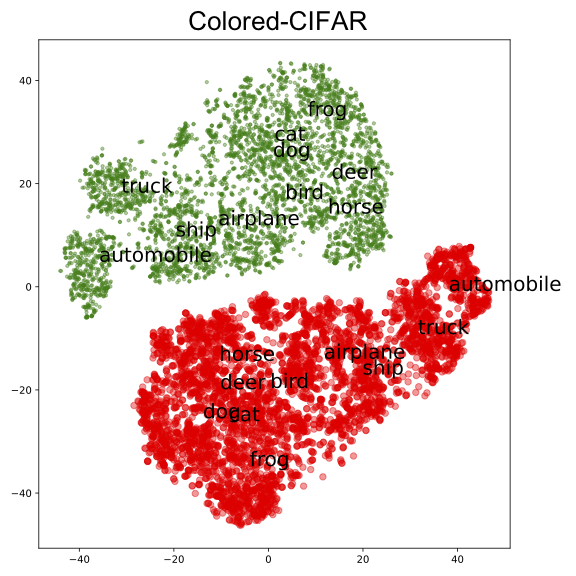}&
    \includegraphics[width=0.24\textwidth]{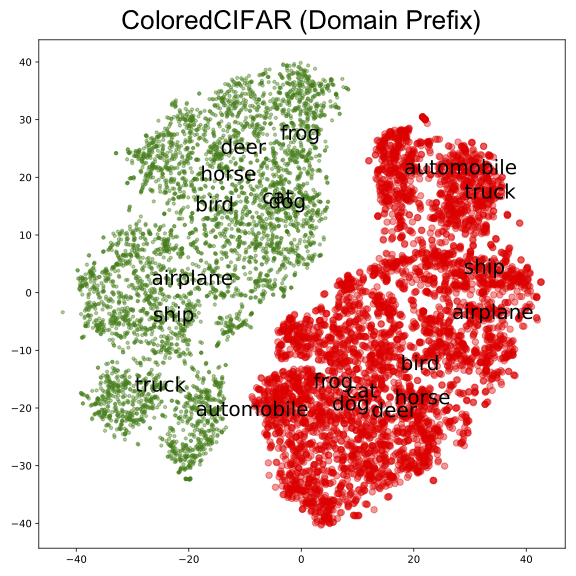}&
    \includegraphics[width=0.24\textwidth]{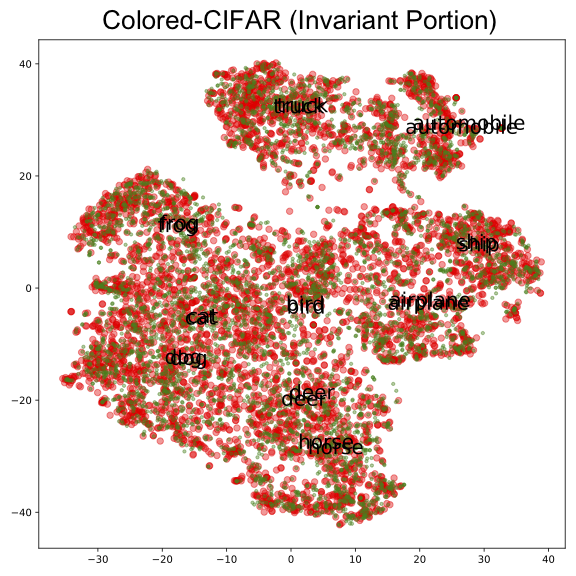}\\
    (a) Accuracy: 90.18 & (b) Accuracy: 78.52 & (c) Accuracy: 87.06 (DDM) & (d) Accuracy: 87.06 (DDM)
    \end{tabular}
    \caption{\textbf{SimCLR Representation t-SNE before and after DDM:} CIFAR-10 representations are naturally clustered by class, however, Colored-CIFAR representations are clustered by domain which leads to a significant reduction in classification performance. When SimCLR is trained with DDM, the prefix alone has domain-distinguishable representations, while the remaining portion of the representation is domain-invariant, clustered by class. This structure notably improves the classification performance.}
    \label{fig:tsne_comp}
\end{figure}

Let us take the example of SimCLR \cite{simclr} trained on CIFAR-10 \cite{cifar10} dataset. In the first t-SNE \cite{tsne} plot in Figure \ref{fig:tsne_comp}(a), we observe that the representations are naturally clustered based on their classes, which allows us to achieve a top-1 accuracy of $90.18$ after linear probing. Let us now define a multi-domain version of CIFAR-10 called \textit{Colored-CIFAR} where, each sample is randomly colored either red or green as shown in the first panel of Figure \ref{fig:most_act_feat}. In this dataset, the domains refer to the colors of the image, while the labels are of the objects. When SimCLR is trained on Colored-CIFAR, there is a significant drop in top-1 accuracy ($78.52$). We observe that the representation space is divided into two large clusters, corresponding to the domains (red or green) as shown in \ref{fig:tsne_comp}(b). We attribute the loss in accuracy to this significant change in representation structure.

We now study the SimCLR representation space of Colored-CIFAR to further understand and explain multi-domain behavior. In Figure \ref{fig:most_act_feat}, in the second panel, we show a heatmap of the domain-wise averaged representations of each class in CIFAR-10. Each column corresponds to specific feature indices of the class-averaged representations. The darker the column, the higher the magnitude of the feature. For fair comparison, we L2 normalize every feature. For ease of visualization, we display only the subset of feature indices (called \textit{most activating features}) that are strongly deviated from the mean in at least one row. The remaining features show low activation across the board and are omitted from visualization \cite{Jing2021UnderstandingDC, kalibhat2022measuring}. Top activating features correspond to important physical attributes discovered from the training data \cite{kalibhat2022measuring, singla2021salient}. Two images of a car, one in each domain, would share all physical attributes except for the color. An ideal self-supervised encoder is expected to encode all physical attributes independent of any domain shift.

However, in multi-domain SimCLR, we observe that there is almost no overlap between the most activating features of each class between the red and green domains. This suggests that the domain information (color) and instance information (actual content of the image) are somewhat interleaved in these representations, causing different sets of features to be strongly activated for the same class based on the domain. In single-domain SimCLR on CIFAR-10 (no colors), the representations only encode content information, which results in linearly separable representations by class. In multi-domain SimCLR on Colored-CIFAR, a combination of both domain and content information is encoded in every representation which directly affects linear classification performance. Therefore, to achieve comparable performance to single-domain setups, we propose to \textbf{disentangle} domain information from representations by plugging in a general-purpose a Domain-Disentanglement Module (DDM) for Self-Supervised Models which is discussed in the next section. 

\section{Domain-Disentanglement Module for Self-Supervised Representations}
As described in the previous section, self-supervised models in their current state, are not trained to learn content and domain information independently. We hypothesize that disentangling domain information from the learned representations can improve the performance of existing state-of-the-art SSL models in multi-domain setups. We therefore propose a general-purpose Domain-Disentanglement Module (DDM) that can be simply attached at any SSL encoder during its pre-training. In this work focus on joint-embedding (involving two transformed views) self-supervised encoders \cite{simclr, swav, mocov2, byol, dino, vicreg, barlowtwins, simsiam} and not masked image models \cite{mae}.

Recall that for a given sample $\bx_i$, its representation is denoted by $f(\bx_i) = \bh_i \in \bbR^r$. Let $y_i$ denote the domain of the $i^{th}$ representation. We allocate the first $k$ features of the representation as the domain prefix, $\bh_{i,0..k}$, denoted by $\bh_{i}^d$ for ease of notation. The remaining portion of the representation $\bh_{i,k..r}$ is denoted by $\bh_{i}^p$. We call $\bh_{i}^d$ as the \textit{domain-variant} portion and $\bh_{i}^p$ as the \textit{domain-invariant} portion. We train the domain prefix of the $i^{th}$ sample according to the following contrastive optimization, $L_{i_{\textit{d\_var}}} = \log \frac{\sum_{j=1}^{2N} \mathbbm{1}_{j \ne i} \mathbbm{1}_{y_i = y_j} sim(\bh_{i}^d, \bh_{j}^d)}{\sum_{j=1}^{2N} \mathbbm{1}_{y_i \ne y_j} sim(\bh_{i}^d, \bh_{j}^d)}$.

where $sim(a, b) = \exp\left({\frac{1}{\tau}\frac{a^Tb}{\|a\|\|b\|}}\right)$. This loss maximizes the similarity of the domain prefixes within each domain and minimizes the similarity of domain prefixes across domains. $\bh_{i}^p$ is learned according to any self-supervised loss like SimCLR, MoCo, DINO etc., denoted by $L_{i_{\textit{ssl}}}$. Splitting the representation in this manner helps us control each portion independently. $L_{\textit{ssl}}$ ensures that all content information is encoded in a self-supervised manner such that representations can be utilized for downstream tasks. $L_{\textit{d\_var}}$ ensures that the domain prefixes across samples of different domains are distinguishable. 

We next ensure that $\bh_{i}^p$ does not contain any domain-related information (domain-invariance constraint). In other words, it should not be possible to predict the the domain label $y_i$ from the representation $\bh_{i}^p$. To achieve this, we pass each $\bh_{i}^p$ through a domain discriminator $D(.)$ and minimize the Wasserstein distance (using the dual form as proposed in \cite{arjovsky2017wasserstein}), $L_{i_{\textit{d\_invar}}} = D(\bh_{i}^p, y_i) - D(\bh_{i}^p, y_{rand})$, where $y_{rand} \sim \mathbbm{P}(y)$, i.e., randomly drawn from the distribution of domain labels. The final optimization for the encoder ($f(.)$) and the discriminator ($D(.)$) is,
\begin{align}
    &\max_{f} \sum_{i = 1}^{2N}  \left[L_{i_{\textit{ssl}}} + \lambda_1 L_{i_{\textit{d\_var}}} + \lambda_2 L_{i_{\textit{d\_invar}}}\right]
\end{align}


where $\lambda_1, \lambda_2$ are tunable hyperparameters. We optimize both the encoder $f(.)$ and the discriminator $D$ using alternating gradient descent ascent. We train $D(.)$ using gradient penalty to improve its stability as proposed in \cite{gulrajani2017improved}. This formulation is similar to \cite{shen2017wasserstein, kattakinda2022invariant}, except that we use Wasserstein Distance to disentangle domain information from the remaining portion of the representation space. In summary, our module DDM consists of splitting the representation space into two parts and applying two additional loss terms, $L_{\textit{d\_var}}$ and $L_{\textit{d\_invar}}$. Note that, DDM can be plugged in while training any existing state-of-the-art self-supervised model. 

In Figure \ref{fig:most_act_feat}, in the last panel, we show the representation space of SimCLR trained on Colored-CIFAR using DDM. We observe that among the most activating features, the first few features (which are part of the domain prefix) are equivalent for all classes within a domain and clearly distinguishable between both domains. The remaining portion of the representation is completely invariant to any domain information as each class shows very similar feature distribution in both red and green domains. In the t-SNE plots (Figure \ref{fig:tsne_comp}(c) and (d)), we observe that the domain prefix is separable by domain whereas the domain-invariant portion shows natural class clusters with overlapping red and green images. This update in structure leads to a significant improvement in top-1 accuracy from 78.52 to 87.06.

\subsection{Experimental Setup} \label{sec:exp_setup}
We use ViT-S \cite{dosovitskiy2021image} as the base encoder ($f(.)$) for all of our experiments. Our domain discriminator ($D(.)$) is an MLP with LeakyReLU activations. The representations are $384$-dimensional with a $24$-dimensional domain prefix. We train the encoder according to various self-supervised baselines including SimCLR \cite{simclr}, MoCo \cite{moco}, BYOL \cite{byol}, DINO \cite{dino}, SimSiam \cite{simsiam} and Barlow Twins \cite{barlowtwins}. We use the same optimization and scheduling for the encoder as the respective papers. While training with DDM, we use the Adam optimizer for the domain discriminator with a learning rate of $0.005$ and cosine-annealing scheduling and $\lambda_1 = \lambda_2 = 0.5$. We experiment with PACS \cite{pacs}, DomainNet \cite{peng2019moment} and the WILDS \cite{koh2021wilds} multi-domain benchmarks. We use Nvidia GeForce RTX A4000 GPUs for pre-training. We evaluate representations using the linear evaluation protocol \cite{8953672, NEURIPS2019_ddf35421, oord2019representation} where we train a linear classifier on top of frozen representations and compute the top-1 accuracy over the training and unseen domains.

\subsection{Self-Supervised Baselines Trained with DDM}
In Figure \ref{fig:single_domain}, we observed that self-supervised baselines (SimCLR, BYOL and MoCo), when trained on a single domain or multiple domains, generalize poorly to unseen domains. These baselines, when pre-trained with DDM, show improved performance on the training domains (PACS Photo, Sketch and Cartoon) as well as significantly improved generalization to the unseen domain (PACS Art Painting). Pre-training on multiple domains with DDM outperforms every self-supervised baseline as shown in Table \ref{tab:pacs} with a maximum of $2.6\%$ improvement on average top-1 accuracy on SimSiam. We also tabulate our results on DomainNet using Painting, Real and Sketch as training domains and Clipart, Infograph and Quickdraw as the unseen domains in Table \ref{tab:domainnet}. We observe that pre-training with DDM improves upon each self-supervised baseline with a maximum of $3.5\%$ improvement on average top-1 accuracy on BYOL. DDM generalizes significantly better than its baselines showing a $6.1\%$ (SimSiam) increase in PACS (Painting) and a $7.4\%$ (DINO) in DomainNet (Clipart).

To further evaluate the generalization of self-supervised baselines with DDM, we utilize the WILDS benchmark \cite{koh2021wilds}. In this benchmark, pre-train on iWildCam (200K samples, 182 classes, 323 domains), Camelyon17 (456K samples, 2 classes, 5 domains), FMoW (141K samples, 62 classes, 80 domains) and RxRx1 (125K samples, 1139 classes, 51 domains). We summarize our results in Table \ref{tab:wilds}. On each benchmark, we observe that DDM outperforms the baselines on the unseen validation set. The accuracy in rxrx1 is low since it is a very hard classification task as it contains 1139 classes and 51 domains. We observe a $5.9\%$ increase linear classification accuracy on iWildCam on SimCLR

\begin{table*}[h]
\caption{SSL baselines trained on PACS (Photo, Sketch and Cartoon) with DDM}
\centering
\resizebox{0.8\textwidth}{!}
{
\begin{tabular}{c|c|c|c|c|c}
\toprule 

\multirow{2}{*}{\textbf{Model}} & \multicolumn{5}{c}{\textbf{Top-1 Accuracy (Baseline / with DDM)}}  \\
& \textbf{Photo} & \textbf{Sketch} & \textbf{Cartoon} & \textbf{Painting (Unseen)} & \textbf{Average} \\ 

\midrule 
\midrule 
SimCLR & 97.54 / \textbf{98.28} & \textbf{98.12} / 97.04 & 98.03 / \textbf{99.24}  & 87.59 / \textbf{89.42} & 95.32 / \textbf{96.00} \\
MoCo & \textbf{93.59} / 93.19 & 92.71 / \textbf{94.36} & 91.63 / \textbf{92.98} & 77.34 / \textbf{78.51} & 88.81 / \textbf{89.76} \\
BYOL & 78.08 / \textbf{81.61} & 76.55 / \textbf{78.24} & 75.55 / \textbf{75.58} & 58.10 / \textbf{62.67} & 72.07 / \textbf{74.53} \\
DINO & 93.67 / \textbf{95.25} & 94.33 / \textbf{96.42} & 79.44 / \textbf{81.77} & 72.12 / \textbf{74.43} & 85.89 / \textbf{86.97} \\
SimSiam &  83.68 / \textbf{84.71} & 80.97 / \textbf{85.44} & \textbf{93.75} / 92.59 & 57.98 / \textbf{64.09} & 79.09 / \textbf{81.71} \\
Barlow Twins & \textbf{85.09} / 83.94 & 85.44 / \textbf{88.07} & 92.0 / \textbf{92.83} & 59.01 / \textbf{62.67} & 80.39 / \textbf{81.89} \\
\bottomrule
\end{tabular}
}
\label{tab:pacs}
\end{table*}
\begin{table}[h]
\caption{SSL baselines trained on DomainNet (Painting, Real and Sketch) with DDM}
\centering
\resizebox{\textwidth}{!}
{
\begin{tabular}{c|c|c|c|c|c|c|c}
\toprule 
\multirow{2}{*}{\textbf{Model}} & \multicolumn{7}{c}{\textbf{Top-1 Accuracy (Baseline / with DDM)}}  \\
& \textbf{Painting} & \textbf{Real} & \textbf{Sketch} & \textbf{Clipart (Unseen)} & \textbf{Infograph (Unseen)} & \textbf{Quickdraw (Unseen)} & \textbf{Average} \\ 
\midrule 
\midrule 
SimCLR & 74.49 / \textbf{75.99}  & 79.31 / \textbf{82.02}  & 85.86 / \textbf{86.26} & 68.60 / \textbf{70.48} & 34.75 / \textbf{39.25} & 22.98 / \textbf{24.38} & 60.99 / \textbf{63.06} \\
MoCo & 70.20 / \textbf{73.08} & \textbf{89.79} / 86.37 & 86.66 / \textbf{88.15} & 65.10 / \textbf{68.91} & 34.56 / \textbf{34.75} & 19.89 / \textbf{22.12} & 61.03 / \textbf{62.23} \\
BYOL & 56.87 / \textbf{59.82} & 77.60 / \textbf{79.67} & 71.43 / \textbf{75.21} & 50.67 / \textbf{55.86} & 27.4 / \textbf{30.68} & 19.33 / \textbf{22.85} & 50.55 / \textbf{54.02} \\
DINO & \textbf{79.53} / 79.11 & 86.46 / \textbf{86.88} & 75.8 / \textbf{76.50} & 66.32 / \textbf{73.76} & 30.83 / \textbf{32.12} & 27.71 / \textbf{29.08} & 61.11 / \textbf{62.90} \\
SimSiam & 77.55 / \textbf{78.78} & 82.02 / \textbf{85.88} & 86.52 / \textbf{88.38} & 67.43 / \textbf{71.53} & 27.03 / \textbf{30.56} & 22.29 / \textbf{25.67} & 60.47 / \textbf{63.47} \\
Barlow Twins & 56.78 / \textbf{61.18} & 79.06 / \textbf{80.16} & 71.56 / \textbf{73.90} & 60.40 / \textbf{64.33} & 26.11 / \textbf{28.82} & 18.67 / \textbf{21.70} & 52.09 / \textbf{55.01} \\
\bottomrule
\end{tabular}
}
\label{tab:domainnet}
\end{table}
\begin{table}[h]
\caption{SSL baselines trained on WILDS with DDM}
    \centering
\resizebox{0.7\textwidth}{!}
{
    \begin{tabular}{c|c|c|c|c}
    \toprule
    & \multicolumn{4}{c}{\textbf{Top-1 Accuracy (Baseline / with DDM)}} \\
    \textbf{Model} & \textbf{iWildCam} & \textbf{Camelyon17}  & \textbf{FMoW} & \textbf{RxRx1} \\
    \midrule
    \midrule
    SimCLR & 66.01 / \textbf{71.87} & 95.19 / \textbf{95.68} & 38.94 / \textbf{41.23} & 8.43 / \textbf{11.20} \\
    MoCo & 67.05 / \textbf{69.12} & 91.45 / \textbf{93.47} & 40.04 / \textbf{40.23} & 5.67 / \textbf{5.93} \\
    BYOL & 71.69 / \textbf{74.88} & 95.15 / \textbf{96.38} & 38.74 / \textbf{39.78} & 4.39 / \textbf{6.20} \\
    DINO & 64.55 / \textbf{68.07} & 94.38 / \textbf{95.38} & 33.57 / \textbf{34.52} & 7.32 / \textbf{7.66} \\
    SimSiam & 60.45 / \textbf{61.16}  & 88.37 / \textbf{89.16} & 39.27 / \textbf{40.05} & 6.39 / \textbf{7.26} \\
    Barlow Twins & 63.17 / \textbf{63.84} & 96.38 / \textbf{97.62} & 44.40 / \textbf{47.46} & 5.79 / \textbf{6.65} \\
    \bottomrule
    \end{tabular}
}
    \label{tab:wilds}
\end{table}


\begin{wrapfigure}[20]{R}{0.45\textwidth}
\centering
   \subfigure{\includegraphics[width=0.22\textwidth]{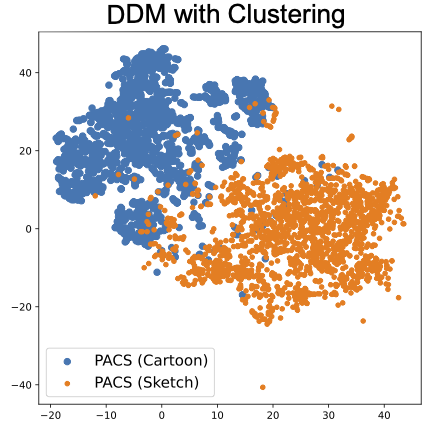}}
    \subfigure{\includegraphics[width=0.22\textwidth]{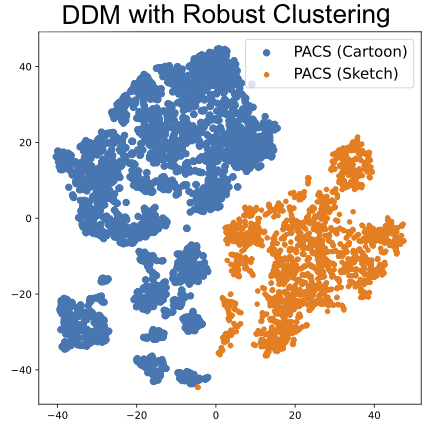}}
  \caption{\textbf{DDM with clustering:} When domain labels are not available, we perform DDM with clustering to identify pseudo-domain-labels. In the above plots we show the t-SNE of the SimCLR representations trained on "Cartoon" and "Sketch" domains in PACS. We observe that DDM with robust clustering produces a better separation between domains.}
  \label{fig:robust_clustering}
\end{wrapfigure}
\section{DDM without Domain Labels}
Most real-world multi-domain datasets are unlabelled (i.e., domain label information is not available). In this section, we develop an extension of DDM for such setups by identifying pseudo domain labels via a clustering approach in the representation space. As it is common in clustering, we assume the number of domains (denoted by $M$) is known. Depending on the multi-domain setup, we can also approximate the number of domains by studying any available meta-data like data sources, geo-location, quality, etc. We can also estimate the number of domains empirically through clustering and visualization.

Domain labels are required in both DDM losses ($L_{\textit{d\_var}}$, $L_{\textit{d\_invar}}$) as described in the previous section. Let us consider a fully unlabelled setup, with no domain labels while the number of domains $M$ is known. We first warm up our self-supervised encoder $f(.)$ treating it as a single-domain setup for a few iterations to get somewhat distinguishable representations by domain. We next cluster the representations into $M$ clusters using K-Means clustering \cite{Hartigan1979}. Using the cluster assignments as pseudo-domain-labels ($y$), we continue training the encoder $f(.)$ along with a discriminator using the DDM optimization, to learn domain-disentangled representations. 

In practice, clustering does not discover $100\%$ accurate domain labels, especially for datasets that are distributionally similar. We therefore use a \textbf{robust clustering} approach coupled with DDM to prevent outlier clustering noise from affecting the pseudo-domain-labels. Suppose we discover $M$ clusters with centroids $\bc_1, \bc_2, \dots, \bc_M$, before assigning pseudo-domain-labels to each sample, we first determine if they are outliers or not. If so, we ignore these samples in the next stages of training to prevent assigning a noisy label to them. We say a representation is {\it not} an outlier if it is significantly closer to one of the clustering centroids compared to another. Concretely, $\bh_i$ is not an outlier if

\begin{align}
    \max \left\{\frac{\|\bh_{i} - \bc_m\|^2}{\|\bh_{i} - \bc_n\|^2}\ : 1 \le m \le M, 1 \le n \le M \right\} > 1 + \epsilon
\label{eq:robust_clustering}
\end{align}

where $\epsilon \ge 0$ is defined as the \textit{outlier threshold}. When $\epsilon$ is high, it means that the given sample is very close to its respective centroid. When $\epsilon$ approaches $0$, it indicates that the sample is almost equidistant from at least two centroids and therefore, may not be reliably assigned one pseudo-label. We ignore such samples going forward in training. When we perform clustering for the first time, we start with $\epsilon = 1$. We repeat the clustering at regular intervals of training on the representations $\bh$ to get improved cluster centroids. Each time we repeat clustering, we decay the value of $\epsilon$ exponentially such that it approaches $0$. By the end of training, all samples will contribute to the training of the self-supervised encoder with DDM. In Figure \ref{fig:robust_clustering}, we illustrate the difference between regular clustering and robust clustering with MDSSL trained on the PACS dataset \cite{pacs} ("Cartoon" and "Sketch" domains). We observe that robust clustering helps in identifying more accurate and distinguishable clusters. 

To evaluate DDM with robust clustering, we combine CIFAR-10 \cite{cifar10}, CIFAR-100 \cite{cifar100} and STL-10 \cite{stl10} to form a multi-domain dataset. The constituent datasets are distributionally similar with several shared classes (CIFAR-10 and STL-10 share 9 out of 10 classes). With this setup, we try to simulate a real-world scenario where data arises from various domains however the actual domains are undefined. We therefore apply DDM with robust-clustering to identify pseudo-domain-labels. We then evaluate the pre-trained representations by linear probing the validation portion of each constituent dataset. We include Tiny-ImageNet \cite{Le2015TinyIV} as an unseen domain to test generalization. 

In Table \ref{tab:cifar10_cifar100_stl10}, we tabulate the results on this prepared multi-domain dataset on various self-supervised baselines with and without DDM and robust clustering. We observe an improvement in the average top-1 accuracy across all baselines with $1.7\%$ improvement in MoCo. DDM shows improved generalization on Tiny-ImageNet with a $2.9\%$ increase in DINO.

\begin{table*}[h]
\centering
\caption{SSL baselines trained on a mixture of CIFAR-10, STL-10 and CIFAR-100 using DDM and robust clustering}
\resizebox{\textwidth}{!}
{
\begin{tabular}{c|c|c|c|c|c}
\toprule 
\multirow{2}{*}{\textbf{Model}} & \multicolumn{5}{c}{\textbf{Top-1 Accuracy (Baseline / with DDM and robust clustering)}}  \\
& \textbf{CIFAR-10} & \textbf{STL-10} & \textbf{CIFAR-100} & \textbf{Tiny-ImageNet (Unseen)} & \textbf{Average} \\ 
\midrule 
\midrule 
SimCLR & 89.43 / \textbf{90.03} & 79.77 / \textbf{81.01} & 63.33 / \textbf{64.90} & 49.58 / \textbf{51.22} & 70.53 / \textbf{71.79} \\
MoCo & \textbf{90.80} / 90.69 & 80.02 / \textbf{81.60} & 61.57 / \textbf{64.28} & 37.16 / \textbf{39.55} & 67.38 / \textbf{69.03} \\
BYOL & 88.31 / \textbf{89.68} & 75.07 / \textbf{75.72} & 64.82 / \textbf{65.56} & 50.04 / \textbf{51.10} & 69.56 / \textbf{70.52} \\
DINO & 90.61 / \textbf{92.96} & \textbf{84.7} / 82.35  & 62.63 / \textbf{63.57} & 49.52 / \textbf{52.46} & 71.87 / \textbf{72.84} \\
SimSiam & 87.02 / \textbf{87.38} & 72.15 / \textbf{73.78} & \textbf{62.08} / 61.90 &  33.11 / \textbf{34.78} & 63.59 / \textbf{64.46} \\
Barlow Twins & 88.31 / \textbf{89.01} & 75.59 / \textbf{76.11} & 65.03 / \textbf{66.89} & 40.27 / \textbf{41.31} & 67.30 / \textbf{68.33} \\
\bottomrule
\end{tabular}
}
\label{tab:cifar10_cifar100_stl10}
\end{table*}

\section{Conclusion}
We proposed a Domain Disentanglement Module (DDM) for self-supervised encoders that provide better structured representations, domain-invariant representations that can be used for diverse multi-domain tasks. DDM also supports training over setups where domain labels are not available by using a robust clustering technique that reduces outlier noise. With DDM, we prevent the need for having to train multiple single-domain encoders and instead  leverage a single encoder to perform comparably on multiple domains. The benefit of invariant representations is better generalization which we show on various benchmarks including PACS, DomainNet and WILDS. 

\section{Acknowledgement}
This project was supported in part by a grant from Capital One, an NSF CAREER AWARD 1942230, ONR YIP award N00014-22-1-2271, ARO’s Early Career Program Award 310902-00001, Meta grant 23010098, HR001119S0026 (GARD), Army Grant No. W911NF2120076, NIST 60NANB20D134, the NSF award CCF2212458 and an Amazon Research Award.

\bibliography{egbib}

\end{document}